\documentclass[11pt]{article}
\usepackage{graphicx}
\usepackage{amssymb}
\usepackage{amscd}
\usepackage{mathrsfs}
\usepackage{longtable,lscape}
\usepackage{amsthm}
\usepackage{amsfonts}
\usepackage{amsmath}
\usepackage{bbm}
\usepackage{float}
\usepackage{url}

\newcommand{\captionfonts}{\footnotesize}
\makeatletter  
\long\def\@makecaption#1#2{%
  \vskip\abovecaptionskip
  \sbox\@tempboxa{{\captionfonts #1: #2}}%
  \ifdim \wd\@tempboxa >\hsize
    {\captionfonts #1: #2\par}
  \else
    \hbox to\hsize{\hfil\box\@tempboxa\hfil}%
  \fi
  \vskip\belowcaptionskip}
\makeatother 
\begin{document}
\title{A New Fundamental Evidence of Non-Classical Structure in the Combination of Natural Concepts}
\author{Diederik Aerts$^1$, Sandro Sozzo$^{2}$ and Tomas Veloz$^{3}$ \vspace{0.5 cm} \\ 
        \normalsize\itshape
        $^1$ Center Leo Apostel for Interdisciplinary Studies, 
         Brussels Free University \\ 
        \normalsize\itshape
         Krijgskundestraat 33, 1160 Brussels, Belgium \\
        \normalsize
        E-Mail: \url{diraerts@vub.ac.be}
          \vspace{0.5 cm} \\ 
        \normalsize\itshape
        $^2$ School of Management and IQSCS, Universit
        y of Leicester \\ 
        \normalsize\itshape
         University Road, LE1 7RH Leicester, United Kingdom \\
        \normalsize
        E-Mail: \url{ss831@le.ac.uk}
          \vspace{0.5 cm} \\ 
        \normalsize\itshape
        $^3$ Department of Mathematics, University of British Columbia, Okanagan Campus \\
        \normalsize\itshape
        3333 University Way, Kelowna, BC Canada V1V 1V7
        \\
        \normalsize\itshape
        and, Instituto de Filosof\'ia y Ciencias de la Complejidad (IFICC)  \\
        \normalsize\itshape
        Los Alerces 3024, \~Nu\~noa, Santiago, Chile  \\
        \normalsize
        E-Mail: \url{tomas.veloz@ubc.ca} \\
              }
\date{}
\maketitle
\begin{abstract}
\noindent
We recently performed  cognitive experiments on conjunctions and negations of two concepts  with the aim of investigating the combination problem of concepts. Our experiments confirmed the deviations (conceptual vagueness, underextension, overextension, etc.)   from the rules of classical (fuzzy) logic and probability theory observed by several scholars in concept theory, while our data were successfully modeled in a quantum-theoretic framework developed by ourselves. In this paper, we isolate a new, very stable and systematic pattern of violation of classicality that occurs  in concept combinations.
In addition, the  strength and regularity of this non-classical effect leads us to 
believe that it occurs at a more fundamental level than the deviations observed up to now.  It is our opinion that we have  identified a deep non-classical mechanism determining not only how concepts are combined but, rather, how they are formed. We show that this effect can be faithfully modeled in a two-sector Fock space structure,  and that it can be exactly explained by assuming that human thought is the supersposition of two processes, a `logical reasoning', guided by `logic', and a `conceptual reasoning' guided by `emergence', and that the latter generally prevails over the former. All these findings provide a new fundamental support to our quantum-theoretic approach to human cognition.
\end{abstract}

\section{Introduction\label{intro}}
An increasing amount of data collected in cognitive psychology reveal that the traditional set-based (Boolean, Bayesian, Kolmogorovian, etc.) models of cognition 
give rise to  fundamentally problematic predictions when applied to human decision making (probability and similarity judgments, concept categorization, natural language processing, behaviour under uncertainty, etc.). In concept theory, these difficulties manifest whenever one tries to apply the prescriptions of classical (fuzzy set) logic and classical probability theory to express the representation of the combination of two or more concepts in terms of the representation of the individual concepts. This problem is known in the literature as the `combination problem', and it was specifically raised by the following findings.

(i) `Pet-fish problem'. Osherson and Smith  considered the `typicality' of specific exemplars with respect to the concepts {\it Pet}, {\it Fish} and their conjunction {\it Pet-Fish} \cite{os1981,os1982}, and  noted that people rate an exemplar such as {\it Guppy} as a very typical example of {\it Pet-Fish}, without rating {\it Guppy} as a very typical example of neither {\it Pet} nor {\it Fish} (`guppy effect'). Therefore, 
the `minimum rule' of fuzzy set conjunction does not hold in the situation of the guppy effect.

(ii) `Overextension and underextension of membership weights'. Hampton measured the `membership weight' of several exemplars with respect to specific pairs of concepts and their conjunction \cite{h1988a} and disjunction \cite{h1988b}, finding systematic deviations from both the minimum and the maximum fuzzy set rule. Adopting an accepted terminology, we say that the membership weight of an exemplar $x$ is `overextended' (`double overextended') with respect to the conjunction `$A \ {\rm and} \ B$' of two concepts $A$ and $B$ if the membership weight of $x$ with respect to `$A \ {\rm and} \ B$' is higher than the membership weight of $x$ with respect to one concept (both concepts) (briefly, $x$ is overextended with respect to the conjunction). Overextension is an example of violation of the minimum fuzzy set rule. Analogously, we say that the membership weight of an exemplar $x$ is `underextended' (`double underextended') with respect to the disjunction `$A \ {\rm or} \ B$' of two concepts $A$ and $B$ if the membership weight of $x$ with respect to `$A \ {\rm or} \ B$' is less than the membership weight of $x$ with respect to one concept (both concepts)  (briefly, $x$ is underextended with respect to the disjunction).

(iii) `Deviations from Boolean logical rules in conceptual negation'. Hampton also measured the membership weights of various exemplars with respect to specific pairs of concepts and their conjunction, e.g., {\it Tools Which Are Also Weapons}, and also conjunction when the second concept is negated, e.g., {\it Tools Which Are Not Weapons} \cite{h1997}. He detected overextension in both types of conjunctions, as well as deviations from classical logical behaviour in conceptual negation. 

(iv) `Sorites-type paradoxes and borderline contradictions'. A borderline contradiction can be formalized as a sentence of the form $P(x) \land \lnot P(x)$, for a vague predicate $P$ and a borderline case $x$, e.g., the sentence ``Mark is rich and Mark is not rich'' \cite{bovw1999}.  Alxatib and Pelletier asked human subjects to estimate the truth value of a sentence such as ``$x$ is tall and not tall'' for a given person $x$ who was showed to the eyes of the subjects \cite{ap2011}. They found that a significant number of subjects estimated such a sentence as true, in particular, for borderline cases.

Difficulties (i)--(iv) entail, in particular, that the formation and combination rules of human concepts do not generally follow the laws of classical (fuzzy set) logic \cite{os1981,os1982,h1988a,h1988b,h1997}. And, more generally, the corresponding experimental data cannot be modeled in a single classical probability space satisfying the axioms of Kolmogorov \cite{ag2005a,ag2005b,a2009a,ags2013,s2014a,s2014b}.

These and other discoveries in experimental psychology led several scholars, including ourselves, to develop more general mathematical formalisms to model cognitive dynamics and processes. Among these alternatives approaches, a very promising one is 
 the `quantum cognition approach', which uses the mathematical formalism of quantum theory
 \cite{ag2005a,ag2005b,a2009a,ags2013,s2014a,s2014b,aa1995,ac2004,vr2004,pb2009,k2010,as2011,bpft2011,bb2012,abgs2013,hk2013,pb2013,wbap2013,as2014,pnas,plosone,IQSA1}.
  Following on our quantum-theoretic approach to human cognition, we recently performed a set of cognitive experiments on conceptual conjunctions of the form `$A$ and $B$', `$A$ and $B'$', `$A'$ and $B$',  and `$A'$ and $B'$', where $A'$ and $B'$ denote the negations of the natural concepts $A$ and $B$, respectively. We identified systematic deviations from classicality  in the form of overextension, as well as double overextension. We thus confirmed the empirical results on concept combinations mentioned above. In addition, we faithfully modeled our experimental data by naturally extending a two-sector Fock space model previously worked out by ourselves \cite{s2014b,asv2015FRONTIERS}.
  
In this paper we present  our recent findings on the identification of quantum structures in the combination of natural concepts. More specifically,  our investigation of the simultaneous representability of conjunctions and negations in Sec. \ref{nonclassicality} leads to five  conditions, which should be satisfied by the data to fit together into one classical probability setting. Analysing the deviations of our data with respect to these five conditions, we find a very strong systematic 
 pattern which is also very stable, in the sense that it gives rise to the same numerical values for the deviation even over the different pairs of concepts that we have experimented on.
  
Inquiring into this regular pattern of violation in Sec. \ref{newpattern}, we show how it constitutes a very strong evidence for the 
presence and dominance of what we have called `conceptual emergence' in our earlier work on Fock space modeling of conceptual combinations \cite{a2009a,ags2013,abgs2013,s2014a,s2014b}. That the violation is numerically the same independent of the considered pair of concepts indicates that we have identified a non-classical mechanism in human thought which is linked to the depth of concept formation itself and to its intrinsic nature, independent of the specific meaning for a specific pair of concepts and a specific set of considered exemplars. 

This was for ourselves the most surprising and unexpected result of the investigation that we present in this paper, and we also consider it as one of the most important of our findings with respect to the modeling of human cognition. In our opinion it inescapably proves that
human thought  does not follow the rules of classical probability, and that this deviation of classical probability is strong and takes place on a deep structural conceptual level.  Let us stress in this respect that this overwhelming systematics of deviation of classicality `could not be detected in the foregoing studies on conjunctions and disjunctions of pairs of concepts,
since these were not considering also negations and their combinations' \cite{h1988a,h1988b,h1997,a2009a,ags2013,s2014a,s2014b}. Indeed, these studies focused on only identifying overextension or underextension, hence they `could not reveal the systematic deviation we detect here, by lack of symmetry'. It was necessary to experiment on conjunction and negation together and derive the five classicality conditions containing the necessary symmetry, to be able to identify this strong and stable pattern of deviation of classicality. 

The second important finding we present in this paper is related to the same identification of the `deviation of classicality pattern' over all the considered pairs of concepts and their negations. Indeed, we not only find a strong and stable numerical deviation independent of the considered concepts and exemplars, but we additionally show in Sec. \ref{explanation} that the numerical size of the deviation is `almost' equal to the deviation of our five classicality conditions `if we substituted the theoretical values for an average quantum model by means of our Fock space model of the situation'. Hence, as a second, equally unexpected and for ourselves surprising result, the data indicate in a very strong way that the deviation is exactly the one that would theoretically be found in case the situation is modelled quantum mechanically by our Fock space model.

Following the above mentioned results we believe our findings to be a strong support for the validity of our quantum-theoretic framework in two-sector Fock space, which we summarize in Sec. \ref{quantum}, and also for our explanatory hypothesis that human reasoning is the superposition of a dominant emergent dynamics and a secondary logical dynamics \cite{IQSA2}. In fact, they indicate that the human mind generally combines the concepts $A$ and the negation $B'$ of the concept $B$ by forming a new concept `$A \ {\rm and} \ B'$', which emerges as an independent conceptual entity. But, superposed to this mechanism, a second dynamics acts, where the new concept `$A \ {\rm and} \ B'$' is recognized as a logical combination of $A$ and $B'$. Our two-sector Fock space is a mathematical framework for representing faithfully these processes, as we conclude in Sec. \ref{conclusions}.

\section{Non-classicality in conceptual conjunctions and negations \label{nonclassicality}}
Our experiments involved 40 human subjects, chosen among our colleagues and friends, and exposed to a `within-subjects design'. This is the typical number of participants in a cognitive experiment estimating conceptual membership -- in particular, Hampton's experiments, which our experiments directly compare with, employed 40 participants \cite{h1988a,h1988b}. We asked the 40 subjects to fill in a questionnaire in which they had to estimate the membership of four sets $C(A,B)$ of exemplars with respect to four pairs $(A,B)$ of natural concepts, and their conjunctions `$A$ and $B$',  `$A$ and $B'$',  `$A'$ and $B$' and `$A'$ and $B'$', where $A'$ and $B'$ denote the negations of the concepts $A$ and $B$, respectively. Hence, for each pair $(A,B)$ of natural concepts, the 40 participants were involved in four subsequent experiments, $e_{AB}$, $e_{AB'}$, $e_{A'B}$ and $e_{A'B'}$, $e_{XY}$ corresponding to the conjunction `$X$ and $Y$', $X=A,A'$, $Y=B,B'$.

For each pair, we considered a set of 24 exemplars and measured their membership with respect to these pairs of concepts and the conjunctions above. For example, for the pair ({\it Fruits}, {\it Vegetables}), we firstly asked the 40 subjects to estimate the membership of 24 exemplars -- {\it Apple}, {\it Parsley}, {\it Olive}, {\it Chili Pepper}, {\it Broccoli}, \ldots, {\it Almond}, \ldots  -- with respect to the concepts {\it Fruits}, {\it Vegetables}, and their conjunction {\it Fruits And Vegetables}. Then, we asked the same 40 subjects to estimate the membership of the same exemplars with respect to the concept  {\it Fruits}, the negation {\it Not Vegetables} of the concept {\it Vegetables}, and their conjunction {\it Fruits And Not Vegetables}. Subsequently, we asked the 40 subjects to estimate the membership of the 24 exemplars with respect to the negation {\it Not Fruits} of the concept  {\it Fruits}, the concept {\it Vegetables}, and their conjunction {\it Not Fruits And Vegetables}. Finally,  we asked the 40 subjects to estimate the membership of the 24 exemplars with respect to the negations {\it Not Fruits},  {\it Not Vegetables}, and their conjunction {\it Not Fruits And Not Vegetables}. Thus, in each experiment, a given subject was asked to provide $24 \cdot 4 \cdot 3 = 288$ judgements, for an overall number of $288 \cdot 4=1152$ judgements.

The choice of the exemplars and pairs of concepts was inspired by the experiments Hampton performed on the disjunction of two concepts \cite{h1988b}. We chose four of the eight pairs of concepts tested by Hampton, together with the same set of exemplars, for each pair. This choice is part of a more general research program in which we want to investigate the validity of specific logical rules, such as de Morgan's laws, in concept combinations.

We estimated conceptual membership by using a `7-point scale'. The subjects were asked to choose a number from the set $\{+3, +2, +1, 0, -1, -2, -3 \}$, where the positive numbers $+1$, $+2$ and $+3$ meant that they considered `the exemplar to be a member of the concept', where $+3$ indicated a strong membership, $+1$ a relatively weak membership. The negative numbers $-1$, $-2$ and $-3$ meant that the subject considered `the exemplar to be a non-member of the concept', where $-3$ indicated a strong non-membership, $-1$ a relatively weak non-membership. We finally converted these `membership or non-membership' estimations into relative frequencies and, in the limit of large numbers, into probabilities, or `membership weights'. More concretely, the membership weight $\mu(A \ {\rm and} \ B)$ is the large number limit of the relative frequency for the exemplar $x$ to be a member of `$A \ {\rm and} \ B$' in the experiment $e_{AB}$, hence the probability $\mu(A \ {\rm and} \ B)$  can be associated with the event `the exemplar $x$ is a member of the concept $A$'. We get the relative frequency by converting the values collected on the 7-point scale and associating a value $+1$ to each positive value on the 7-point scale,  $0$ to each negative number, and $0.5$ to each $0$ on the same 7-point scale.

We reported the collected data in \cite{asv2015FRONTIERS}, together with the analysis of their statistical significance. We found that most of these data `deviate from classicality', that is, they cannot generally be modeled by representing conceptual conjunctions and negations with the usual connectives of classical (fuzzy set) logic. We therefore confirmed the results found by other scholars on the combinations of two concepts \cite{os1981,h1988a,h1988b,h1997,ap2011} (see Sec. \ref{intro}). More specifically, we identified `overextension' in the conjunction $\mu(A \ {\rm and} \ B)$, e.g., the exemplar {\it Apple} scores $\mu(A)=1$ with respect to the concept {\it Fruits}, $\mu(B)=0.23$ with respect to the concept {\it Vegetables}, and $\mu(A \ {\rm and} \ B)=0.6$ with respect to the conjunction {\it Fruits And Vegetables}. We also identified `overextension when one concept is negated', e.g., {\it Pepper} scores 0.99 with respect to {\it Spices},  0.58 with respect to {\it Not Herbs}, and 0.9 with respect to {\it Spices and Not Herbs}, while the membership weights of {\it Desk} with respect to {\it Not Home Furnishing}, {\it Furniture} and {\it Not Home Furnishing And Furniture} is 0.31, 0.95 and 0.75, respectively. More, we detected `overextension when both concepts are negated', e.g., we have $\mu(A')=0.12$, $\mu(B')=0.81$ and $\mu(A' \ {\rm and} \ B')=0.43$ for {\it Goldfish}, with respect to {\it Not Pets} and {\it Not Farmyard Animals} and {\it Not Pets And Not Farmyard Animals}, respectively. Finally, `double overextension' is present in various cases, e.g. the membership weight of {\it Olive} with respect to {\it Fruits And Vegetables} is 0.65, which is greater than both 0.53 and 0.63, i.e. the membership weights of {\it Olive} with respect to {\it Fruits} and {\it Vegetables}. {\it Prize Bull} scores 0.13 with respect to {\it Pets} and 0.26 with respect to {\it Not Farmyard Animals}, but its membership weight with respect to {\it Pets And Not Farmayard Animals} is 0.28. Also, {\it Door Bell} gives 0.32 with respect to {\it Not Home Furnishing} and 0.33 with respect to {\it Furniture}, while it gives 0.34 with respect to {\it Not Home Furnishing And Furniture}.

More generally, the collected membership weights violate the rules of classical probability. To systematically identify such deviations from classicality we proved the following theorem providing a characterization of the representability of this set of experimental data in a classical probability space \cite{asv2015FRONTIERS}.

\bigskip
\noindent 
{\bf Theorem.} {\it If the membership weights $\mu(A), \mu(B), \mu(A'), \mu(B')$, $\mu(A\ {\rm and}\ B)$, $\mu(A\ {\rm and}\ B')$, $\mu(A'\ {\rm and}\ B)$ and $\mu(A'\ {\rm and}\ B')$ of an exemplar $x$ with respect to the concepts $A$, $B$, $A'$ and $B'$ and the conjunctions `$A$ and $B$', `$A$ and $B'$', `$A'$ and $B$' and `$A'$ and $B'$' are all contained in the interval $[0,1]$, they are classical conjunction data if and only if they satisfy the following conditions.}
\begin{eqnarray} \label{condbis01}
&\mu(A)=\mu(A\ {\rm and}\ B)+\mu(A \ {\rm and}\ B') \\ \label{condbis02}
&\mu(B)=\mu(A\ {\rm and}\ B)+\mu(A' \ {\rm and}\ B) \\ \label{condbis03}
&\mu(A')=\mu(A'\ {\rm and}\ B')+\mu(A' \ {\rm and}\ B) \\ \label{condbis04}
&\mu(B')=\mu(A'\ {\rm and}\ B')+\mu(A \ {\rm and}\ B') \\ \label{condbis05}
&\mu(A\ {\rm and}\ B)+\mu(A\ {\rm and}\ B')+\mu(A'\ {\rm and}\ B)+\mu(A'\ {\rm and}\ B')=1
\end{eqnarray}

\bigskip
\noindent
The theorem stated above provides the most symmetric conditions for simultaneously representing the data in the experiments $e_{XY}$, $X=A,A'$, $Y=B,B'$ in a single classical space satisfying the axioms of Kolmogorov \cite{k1950,a1986,p1989}. One recognizes at once the marginal law of classical probability in Eqs. (\ref{condbis01})--(\ref{condbis04}), while Eq. (\ref{condbis05}) expresses normalization of probabilities. If we introduce the functions
\begin{eqnarray}
&I_{A}=\mu(A)-\mu(A\ {\rm and}\ B)-\mu(A \ {\rm and}\ B') \label{negationA} \\
&I_{B}=\mu(B)-\mu(A\ {\rm and}\ B)-\mu(A' \ {\rm and}\ B) \label{negationB}\\
&I_{A'}=\mu(A')-\mu(A'\ {\rm and}\ B')-\mu(A' \ {\rm and}\ B) \label{negationA'}\\
&I_{B'}=\mu(B')-\mu(A'\ {\rm and}\ B')-\mu(A \ {\rm and}\ B') \label{negationB'} \\
&I_{ABA'B'}=1-\mu(A\ {\rm and}\ B)-\mu(A\ {\rm and}\ B')-\mu(A'\ {\rm and}\ B)-\mu(A'\ {\rm and}\ B') \label{normalization}
\end{eqnarray}
then the theorem above can be restated by saying that the membership weights $\mu(X)$, $\mu(Y)$, $\mu(X \ {\rm and}\ Y)$ are classical conjunction data if and only if $I_{X}=I_{Y}=I_{ABA'B'}=0$, $X=A,A'$, $Y=B,B'$. These functions significantly deviate from 0 in our data --  {\it Field Mouse} has $I_{ABA'B'}=-0.46$, while {\it Chili Pepper} has $I_{A}=-0.73$ and {\it Pumpkin} has $I_{B'}=-0.13$ -- which we proved by means a t-student test for paired samples for means (with Bonferroni correction) \cite{asv2015FRONTIERS}.

\section{A new systematic and stable pattern of violation\label{newpattern}}
The analysis in Sec. \ref{nonclassicality} follows on the lines traced in \cite{a2009a,ags2013,s2014a,s2014b} with respect to the identification of non-classical effects in concept combination, such as, conceptual vagueness, guppy-type effects, overextension and underextension of membership weights, and so on. However, the symmetry of the cognitive experiments we performed, together with the completeness of the data we collected in these experiments \cite{asv2015FRONTIERS}, allow us to identify a different, unexpected, but even more fundamental, non-classical effect in the combination of conceptual conjunction and negation. We study this effect in this section.

The values of the functions $I_{A}$, $I_{B}$, $I_{A'}$, $I_{B'}$ and $I_{ABA'B'}$ for the different pairs of concepts are reported in Appendix \ref{tables}, Tabs. 1--4. By pure inspection of Tabs. 1--4, one can immediately recognize that some systematicity is involved, namely, the terms $I_{A}$, $I_{B}$, $I_{A'}$, $I_{B'}$ and $I_{ABA'B'}$ in Eqs. (\ref{negationA})--(\ref{normalization}) show similar patterns, since they are characterized by approximately constant numerical values. 
Numerically, one has that the mean value of $I_A$ is -0.42, across all exemplars, and its standard deviation is 0.09. $I_B$ has mean value -0.43 and standard deviation 0.075. $I_A'$ has mean value -0.35 and standard deviation 0.09. $I_B'$ has mean value -0.33 and standard deviation 0.09. Finally, $I_{ABA'B'}$ has mean value -0.81 and standard deviation 0.13. But, a further analysis reveals that this pattern of violation exhibits specific features:

(i) it cannot be explained by means of traditional classical probabilistic approaches, since we should have $I_{X}=I_{Y}=I_{ABA'B'}=0$, $X=A,A'$, $Y=B,B'$ in that case, as we have seen in Sec. \ref{nonclassicality}.

(ii) it is `highly stable', in the sense that the functions $I_{A}$, $I_{B}$, $I_{A'}$, $I_{B'}$ and $I_{ABA'B'}$ are very likely between -1 and 0;

(iii) is is `systematic', in the sense that the values of $I_{A}$, $I_{B}$, $I_{A'}$, $I_{B'}$ and $I_{ABA'B'}$ are approximately the same
 across all exemplars;

(iv) it is `regular', in the sense that the functions $I_{A}$, $I_{B}$, $I_{A'}$, $I_{B'}$ and $I_{ABA'B'}$ are independent of the pair of concepts that are considered.
		
Observations (i)--(iv) led us to suspect that $I_{X}$, $I_{Y}$, $X=A,A',Y=B,B'$ and $I_{ABA'B'}$ are indeed constant functions across all exemplars and pairs of concepts. To this end we firstly performed a `linear regression analysis' of our data, sorted from smaller to larger, so we could check whether these quantities can be represented by a straight-line of the form $y=mx+q$, with $m=0$. This was indeed the case.  For $I_{A}$, we obtained $m=3.0 \cdot 10^{-3}$ with $R^{2}=0.94$; for $I_{B}$, we obtained $m=2.9 \cdot 10^{-3}$ with $R^{2}=0.93$; for $I_{A'}$, we obtained $m=2.6 \cdot 10^{-3}$ with $R^{2}=0.96$; for $I_{B'}$, we obtained $m=3.1 \cdot 10^{-3}$ with $R^{2}=0.98$; for $I_{ABA'B'}$, we obtained $m=4 \cdot 10^{-3}$ with $R^{2}=0.92$. Successively, we computed the $95\%$-confidence interval for these parameters and obtained interval $(-0.51, -0.33)$ for $I_A$, interval $(-0.42, -0.28)$ for $I_{A'}$, interval $(-0.52, -0.34)$ for $I_B$, interval $(-0.40, -0.26)$ for $I_{B'}$, and interval $(-0.97, -0.64)$ for $I_{ABA'B'}$. We can thus conclude that the measured parameters systematically fall within a narrow band centered at very similar values. The very high values of $R^2$ are enough to claim that the non-classical effect we have identified here is statistically significant and not due to random errors. It is however worth to make a subtler and very interesting distinction between the linear regression analysis of $I_{X}$, $I_{Y}$, $X=A,A',Y=B,B'$, and $I_{ABA'B'}$. Indeed, we calculated the p-values for the corresponding regression analysis. We found that $I_A$ and $I_B$ give high p-values, 0.31 and 0.32, respectively, thus confirming the non-significance of the regression analysis and the independence of the values of $I_A$ and $I_B$ from the exemplar and pair of concepts. On the contrary, the regression analysis of $I_{A'}$ and $I_{B'}$ (and $I_{ABA'B'}$) gives low p-values ($p<0.05$), hence one cannot exclude that these functions have a weak dependence on the exemplar that is considered. However, the values of the coefficients of the linear regression is so small that it can be neglected, thus confirming the fact that $I_A$, $I_B$, $I_{A'}$, $I_{B'}$, $I_{ABA'B'}$ are constant functions across exemplars and pairs of concepts.

We also calculated the correlation matrix for the values of $I_{X}$, $I_{Y}$, $X=A,A',Y=B,B'$ and $I_{ABA'B'}$, finding that $I_X$ and $I_Y$ positively related each other and with $I_{ABA'B'}$, $X=A,A'$, $Y=B,B'$, while $I_A$ has a weak negative relationship with $I_{A'}$ and $I_B$ has a weak negative relationship with $I_{B'}$, as expected. The correlation matrix is reported in Tab. 5.

The pattern above is incredibly stable, systematic and regular, because it is independent on the specific exemplar, the pair of concepts and the type of conjunction that are considered. This result is really surprising for our research, since it was completely unexpected a priori. As we have anticipated in Sec. \ref{intro}, this non-classical effect could not have been isolated in absence of a very symmetric cognitive test, where all conjunctions and negations are tested together. This is probably why it was not identified in traditional cognitive experiments on conceptual combinations \cite{h1988a,h1988b,h1997,s2014b}, which instead aimed at revealing overextension and underextension, hence they lacked the required symmetry on purpose. And we believe that this deviation from classical logical and probabilistic rules occurs at a deeper, more fundamental, level than the known deviations due to overextension and underextension, as mentioned in Sec. \ref{intro}. In our opinion, it expresses a fundamental mechanism through which concepts are formed.  Therefore, we can briefly say that the traditional overextension and underextension 
reveal aspects  mainly related to concept combination, but this non-classical effect captures aspects related to concept formation in its foundation, independent of meaning. This aspect can be supported with a further statistical test. Indeed, we computed the p-values corresponding to a t-test for paired samples for means for $I_X$ and $I_Y$, $X=A,A'$, $Y=B,B'$. We found that $I_A$ versus $I_B$ gives a p-value 0.85, and 
$I_A'$ versus $I_B'$ gives a p-value 0.08, hence it is very reasonable to maintain that the difference between $I_A$ and $I_B$ and between $I_A'$ and $I_B'$ is due to chance. On the contrary, $I_A$ versus $I_A'$ gives a p-value $3.47EE-9$, $I_A$ versus $I_B'$ gives a p-value $3.25EE-15$, $I_B$ versus $I_A'$ gives a p-value $2.44EE-15$, and $I_B$ versus $I_B'$ gives a p-value $1.69EE-13$. Hence, it is very reasonable to maintain that the differences between $I_A$ and $I_A'$,  between $I_A$ and $I_B'$,   between $I_B$ and $I_A'$, and between $I_B$ and $I_B'$ are not due to chance.  What can we infer from this result? Probably, it is the fact that a concept or its negation is measured that makes a difference in the values of these functions, which again supports our claim that the non-classical mechanism we have identified is more due to conceptual formation than to conceptual combination.

The still much higher p-value of $I_A$ versus $I_B$ (0.85) as compared to the p-value of $I_{A'}$ versus $I_{B'}$ (0.08) could be due to $A$ and $B$ being `completely finished and stable in their formation as a concept', while $A'$ and $B'$ being `more momentaneous and hence fragile in their concept formation' -- we plan to investigate these aspects further in the future. For example, if $A$ and $B$ were taken to be `adjectives' rather than `subjectives' as concepts, their negations might be less fragile.
   
The results attained here could already be considered as crucial for claiming that the violation of classicality occurs at a deep structural conceptual level, but this is not the end of the story.  We will see in Sec. \ref{explanation} that the stability of this violation can exactly be explained in a two-sector Fock space quantum framework elaborated by ourselves to model conjunctions and disjunctions of two concepts, and recently extended to also incorporate conceptual negation.

\section{A quantum-theoretic modeling for conjunctions and negations\label{quantum}}
We sketch in this section the quantum-theoretic modeling we developed to represent our experimental data on the conjunction and the negation of two concepts. Though we introduced some conceptual and technical novelties in it, our quantum model exhibits the same general features of the model originally elaborated to represent the conjunction and the disjunction of two concepts, of which the present model is a natural extension. For the sake of brevity, we will omit presenting proofs and technical  steps. The interested reader can refer to \cite{asv2015FRONTIERS}.

The quantum-theoretic framework where we represent conceptual entities is mathematically described by a Fock space, i.e. an infinite direct sum of Hilbert spaces. In the simplest case of two combining concepts, it is however sufficient to consider a two-sector Fock space ${\cal F}={\cal H}\oplus({\cal H}\otimes{\cal H})$, i.e. the direct sum $\oplus$ of two Hilbert spaces, an individual Hilbert space $\cal H$ and the tensor product ${\cal H} \otimes {\cal H}$ of two isomorphic copies of $\cal H$. The Hilbert space ${\cal H}$ is called the `sector 1 of $\cal F$', the tensor product Hilbert space ${\cal H}\otimes{\cal H}$ is called the `sector 2 of $\cal F$'.

Let us now come to the representation of the conjunctions `$A$ and $B$', `$A$ and $B'$', `$A'$ and $B$' and `$A'$ and $B'$' and the negations $A'$ and $B'$ of two concepts $A$ and $B$. We represent the concepts $A$, $B$, $A'$ and $B'$ by the unit vectors  $|A\rangle$, $|B\rangle$, $|A'\rangle$ and $|B'\rangle$, respectively, of the Hilbert space $\cal H$, and assume that $\{   |A\rangle, |B\rangle, |A'\rangle, |B'\rangle \}$ is an orthonormal (ON) set in $\cal H$. Let $x$ be an exemplar and, for every $X=A,A'$, $Y=B,B'$,
  let $\mu(X)$ and $\mu(Y)$ be the membership weights of $x$ with respect to the concepts $X$ and $Y$, respectively, and let $\mu(X \ {\rm and} \ Y)$ denote the membership weight of $x$ with respect to the conjunction `$X \ {\rm and} \ Y$'. We represent the `yes-no' decision measurement of a subject who estimates the membership of the exemplar $x$ with respect to the concept $X$ or $Y$ by the orthogonal projection operator $M$ over $\cal H$ (for the yes outcome, by the orthogonal complement $\mathbbmss{1}-M$ for the no outcome). By applying standard quantum probabilistic rules we have, for every $X=A,A'$ and $Y=B,B'$,  $\mu(X)=\langle X|M|X\rangle$ and $\mu(Y)=\langle Y|M|Y\rangle$. 

For the sake of simplicity, we now split the two-sector Fock space representation in a `sector 1 reresentation' and a `sector 2 representation'.

Let us firstly consider `sector 1 representation'. For every $X=A,A'$, $Y=B,B'$, the conjunction `$X \ {\rm and} \ Y$' of the concepts $X$ and $Y$ is represented by the normalized superposition vector $\frac{1}{\sqrt{2}}(|X\rangle+|Y\rangle)$  in sector 1. Therefore, sector 1 representation models conceptual combinations as  the emergence of a new concept, namely the `conjunction concept'. The `yes-no' decision measurement of a subject who estimates the membership of the exemplar $x$ with respect to the conjunction `$X \ {\rm and} \ Y$'  is represented by the orthogonal projection operator $M$ over $\cal H$. By applying standard quantum probabilistic rules we have that, for every $X=A,A'$ and $Y=B,B'$, the membership weight of $x$ with respect to the conjunction `$X \ {\rm and} \ Y$' is given by
\begin{equation}
\label{BornRuleSect1}
\mu(X \ {\rm and} \ Y)=\frac{1}{2}(\langle X|+\langle Y|)M(|X\rangle+|Y\rangle)=\frac{1}{2}(\mu(X)+\mu(Y))+\Re \langle X|M|Y\rangle
\end{equation}
in sector 1, where $\Re \langle X|M|Y\rangle$ is the real part of the complex number $\langle X|M|Y\rangle$ and is called the `interference term' by analogy with the traditional interference of quantum physics \cite{d1958}.

Let us now come to `sector 2 representation'. For every $X=A,A'$, $Y=B,B'$, the conjunction `$X \ {\rm and} \ Y$' of the concepts $X$ and $Y$ is represented by the (generally) entangled unit vector $|C\rangle$ of ${\cal H}\otimes {\cal H}$ in sector 2. The `yes-no' decision measurement of a subject who estimates the membership of the exemplar $x$ with respect to the conjunction `$X \ {\rm and} \ Y$' is represented by the orthogonal projection operator $M_{X}\otimes M_{Y}$ over ${\cal H}\otimes {\cal H}$, $X=A,A'$, $Y=B,B'$, where $M_{A}=M=M_{B}$ and $M_{A'}=\mathbbmss{1}-M=M_{B'}$, in sector 2. Therefore, sector 2 representation models conceptual combinations by applying the rules of logic -- e.g.,  the decision about memberhip of the negation $A'$ is represented by the orthogonal complement ${\mathbbmss 1}-M$ of $M$, where $M$ corresponds to the membership of the positive concept $A$ -- though in a probabilistic form. By applying standard quantum probabilistic rules we have that, for every $X=A,A'$ and $Y=B,B'$, the membership weight of $x$ with respect to the conjunction `$X \ {\rm and} \ Y$' is given by
\begin{equation}
\label{BornRuleSect2}
\mu(X \ {\rm and} \ Y)=\langle C| M_{X}\otimes M_{Y}|C\rangle
\end{equation}
in sector 2.

Let us now consider the overall representation in the two-sector Fock space ${\cal F}={\cal H}\oplus({\cal H}\otimes{\cal H})$. For every $X=A,A'$, $Y=B,B'$, the conjunction `$X \ {\rm and} \ Y$' of the concepts $X$ and $Y$ is represented by the normalized superposition vector
\begin{equation} \label{fockstateXY}
\Psi(X,Y)=m_{XY}e^{i\theta}|C\rangle + n_{XY} e^{i\rho} \frac{1}{\sqrt{2}}(|X\rangle + |Y\rangle)
\end{equation}
where $m_{XY}^2+n_{XY}^2=1$. The `yes-no' decision measurement of a subject who estimates the membership of the exemplar $x$ with respect to the conjunction `$X \ {\rm and} \ Y$' is represented by the orthogonal projection operator $(M_{X}\otimes M_{Y}) \oplus M$ over ${\cal F}$, $X=A,A'$, $Y=B,B'$, where $M_{A}=M=M_{B}$ and $M_{A'}=\mathbbmss{1}-M=M_{B'}$. Again by applying standard quantum probabilistic rules we have that, for every $X=A,A'$ and $Y=B,B'$, the membership weight of $x$ with respect to the conjunction `$X \ {\rm and} \ Y$' is given by
\begin{eqnarray} \label{BornRuleFock}
\mu(X \ {\rm and} \ Y)&=&\langle \Psi(X,Y)|(M_{X}\otimes M_{Y}) \oplus M|\Psi(X,Y)\rangle= \nonumber \\
&=&m_{XY}^{2}\langle C| M_{X}\otimes M_{Y}|C\rangle+n_{XY}^{2} \Big ( \frac{1}{2}(\mu(X)+\mu(Y))+\Re \langle X|M|Y\rangle \Big )
\end{eqnarray}
We proved in \cite{asv2015FRONTIERS} that a general two-sector Fock space modeling for our set of experimental data can be constructed over the complex Hilbert space $\mathbb{C}^{8}$. Indeed, let us denote by $\{ |i\rangle \}_{i=1,\ldots,8}$ the canonical basis of $\mathbb{C}^{8}$ and, for every $X=A,A'$ and $Y=B,B'$, we choose $|X\rangle=e^{i \phi_{X}}\sum_{i=1}^{8}x_i|i\rangle$ and $|Y\rangle=e^{i \phi_{Y}}\sum_{i=1}^{8}y_i|i\rangle$ in this canonical basis. We also choose $M=\sum_{i=5}^{8}|i\rangle\langle i|$, thus $\mathbbmss{1}-M=\sum_{i=1}^{4}|i\rangle\langle i|$, and write $|C\rangle=\sum_{i,j=1}^{8}c_{ij}e^{i\gamma_{ij}}|i\rangle\otimes|j\rangle$. By making the required calculations, one can show that, for every $X=A,A'$ and $Y=B,B'$, Eq. (\ref{BornRuleFock}) takes the form
\begin{equation} \label{BornRuleFockC8}
\mu(X \ {\rm and} \ Y)=m_{XY}^{2}\alpha_{XY}+n_{XY}^{2} \Big ( \frac{1}{2}(\mu(X)+\mu(Y))+\beta_{XY} \cos\phi_{XY} \Big )
\end{equation}
where $m_{XY}^2+n_{XY}^2=1$, $\alpha_{XY}=\alpha_{XY}(c_{ij})$, $0 \le \alpha_{XY} \le 1$, $\beta_{XY}=\beta_{XY}(x_i,y_i)$, $-1\le \beta_{XY}\le 1$, and $\cos\phi_{XY}=\phi_X-\phi_Y$ \cite{asv2015FRONTIERS}.

Almost all the experimental data we collected can be represented in our two-sector Fock space model. For the sake of brevity, we only report the representation for the cases that were double overextended, hence classically highly problematical.

{\it Olive}.				
$m_{AB}^{2}=0.18$, $n_{AB}^{2}=0.82$, $\alpha_{AB}=0.19$, $\beta_{AB}=0.31$, $\phi_{AB}=57.31^{\circ}$, 
$m_{AB'}^{2}=1$, $n_{AB'}^{2}=0$, $\alpha_{AB'}=0.34$, $\beta_{AB'}=0.35$, $\phi_{AB'}=95.32^{\circ}$,
$m_{A'B}^{2}=0.6$, $n_{A'B}^{2}=0.4$, $\alpha_{A'B}=0.44$, $\beta_{A'B}=-0.33$, $\phi_{A'B}=103.43^{\circ}$,
$m_{A'B'}^2=0.27$,  $n_{A'B'}^{2}=0.73$,  $\alpha_{A'B'}=0.03$, $\beta_{A'B'}=0.35$, $\phi_{A'B'}=85.56^{\circ}$.

{\it Prize Bull}.
$m_{AB}^{2}=0.22$, $n_{AB}^{2}=0.78$, $\alpha_{AB}=0.06$, $\beta_{AB}=-0.29$, $\phi_{AB}=105.71^{\circ}$,
$m_{AB'}^{2}=0.17$, $n_{AB'}^{2}=0.83$, $\alpha_{AB'}=0.07$, $\beta_{AB'}=0.16$, $\phi_{AB'}=40.23^{\circ}$,
$m_{A'B}^{2}=0.29$, $n_{A'B}^{2}=0.71$, $\alpha_{A'B}=0.7$, $\beta_{A'B}=-0.14$, $\phi_{A'B}=111.25^{\circ}$,
$m_{A'B'}^2=0.27$, $n_{A'B'}^{2}=0.73$, $\alpha_{A'B'}=0.16$, $\beta_{A'B'}=-0.28$, $\phi_{A'B'}=52.51^{\circ}$.

{\it Door Bell}.
$m_{AB}^{2}=0.23$, $n_{AB}^{2}=0.77$, $\alpha_{AB}=0.12$, $\beta_{AB}=-0.32$, $\phi_{AB}=102.81^{\circ}$,
$m_{AB'}^{2}=0.83$, $n_{AB'}^{2}=0.17$, $\alpha_{AB'}=0.63$, $\beta_{AB'}=0.18$, $\phi_{AB'}=117.67^{\circ}$,
$m_{A'B}^{2}=0.42$, $n_{A'B}^{2}=0.58$, $\alpha_{A'B}=0.21$, $\beta_{A'B}=0.27$, $\phi_{A'B}=67.37^{\circ}$,
$m_{A'B'}^2=0.18$, $n_{A'B'}^{2}=0.82$, $\alpha_{A'B'}=0.04$, $\beta_{A'B'}=0.31$, $\phi_{A'B'}=77.65^{\circ}$.

The interested reader can find a complete representation in \cite{asv2015FRONTIERS}, together with further technical details on our quantum-theoretic framework. We instead discuss here two important aspects of our quantum-theoretic modeling, namely, the connections between modeling parameters and data points, and some intuitive reasons for this choice of the concepts representation.

Let us preliminarily observe that, for every $X=A,A'$, $Y=B,B'$, there are three data points, $\mu(X)$, $\mu(Y)$ and $\mu(X \ {\rm and} \ Y)$, which are connected with the modeling parameters $\alpha_{XY}$, $\beta_{XY}$, $m^{2}_{XY}$, $n^{2}_{XY}=1-m^{2}_{XY}$ and $\cos\phi_{XY}$ by Eq. (\ref{BornRuleFockC8}). The parameters are bounded by the following conditions:
\begin{eqnarray}
& -1 \le \beta_{XY} \le 1 \quad 0 \le \alpha_{XY} \le 1 \\ 
&\sum_{X=A,A}\sum_{Y=B,B'} \alpha_{XY}=1 \\
&\phi_{XY}=\arccos \frac{\Big \{  \mu(X \ {\rm and} \ Y)-\frac{1}{2}(\mu(X)+\mu(Y))-m^{2}_{XY} [\alpha_{XY}-\frac{1}{2}(\mu(X)+\mu(Y))] \Big \}}{(1-m^{2}_{XY})\beta_{XY}}
\end{eqnarray}
Secondly, let us discuss more extensively the reasons why we represent a conceptual conjunction by means of a superposed state vector in sector 1 of Fock space and by means of an entangled state vector in sector 2. For every $X=A,A'$, $Y=B,B'$, the superposition state vector $\frac{1}{\sqrt{2}}(|X\rangle+|Y\rangle)$  represents the conjunction `$X \ {\rm and} \ Y$' in sector 1. This choice expresses the fact that sector 1 of Fock space formalizes the `genuinely emergent aspects of conceptual conjunctions', -- $\frac{1}{\sqrt{2}}(|X\rangle+|Y\rangle)$  is a new state vector, obtained from the state vectors $|X\rangle$ and $|Y\rangle$ but without requiring any logical rule being satisfied. This will be evident in Sec. \ref{explanation} after introducing `quantum emergent thought'. The choice of representing the conjunction `$X$ and $Y$'  by an entangled state vector $|C\rangle$ in sector 2 allows instead the possibility of describing events
 that are not statistically independent in this sector, -- if we represent `$X$ and $Y$'  by the product state vector $|X\rangle\otimes |Y\rangle$ in sector 2, this leads to a probability $\mu(X)\mu(Y)$ in this sector. In addition, there is a striking connection between logic and classical probability when conjunction and negation of concepts are considered together. Namely, the logical probabilistic structure of sector 2 of Fock space sets the limits of classical probabilistic models, and  vice versa. In other words, the experimentally collected membership weights $\mu(X)$, $\mu(Y)$ and $\mu(X \ {\rm and}\ Y)$, $X=A,A'$, $Y=B,B'$, 
 satisfy Eqs. (\ref{condbis01})--(\ref{condbis05}), if and only if an entangled state vector $|C\rangle$ and a decision measurement projection operator $M$ exist such that $\mu(X)$, $\mu(Y)$ and $\mu(X \ {\rm and}\ Y)$, $X=A,A'$, $Y=B,B'$, 
 can be represented in sector 2 of Fock space (see \cite{asv2015FRONTIERS} for the proof). This will be evident in Sec. \ref{explanation} after introducing `quantum logical thought'.

\section{Explaining the observed patterns\label{explanation}}
Our experimental data on conjunctions and negations of natural concepts confirm that classical probability does not generally work when humans combine concepts, as we have seen in Sec. \ref{nonclassicality}. However, we have also proved in Sec. \ref{newpattern} that the deviations from classicality cannot be reduced to overextension and underextension, while our quantum-theoretic framework in two-sector Fock space has received remarkable confirmation, as we have seen in Sec. \ref{quantum}. It is now worth to provide an explanation for the validity of our quantum cognition approach.

We recently put forward an explanatory hypothesis with respect to the deviations from classical logical reasoning that were observed in human cognition \cite{a2009a,ags2013,IQSA2}. According to our explanatory hypothesis, human reasoning is a specifically structured superposition of two processes, a `logical reasoning' and an `emergent reasoning'. The former `logical reasoning' combines cognitive entities, such as concepts, combinations of concepts, or propositions,  by applying the rules of logic, though 
generally in a probabilistic form. The latter `emergent reasoning' enables formation of combined cognitive entities as 
newly emerging entities, in the case of concepts, new concepts, in the case of propositions, new propositions, 
carrying new meaning, linked to the meaning of the constituent cognitive entities, but with a linkage not defined by the algebra of logic. The two mechanisms act simultaneously and in superposition in human thought during a reasoning process, the first one is guided by an algebra of `logic', the second one follows a mechanism of `emergence'. And, more, human reasoning can be exactly  mathematically formalized in a two-sector Fock space. More specifically, sector 1 of Fock space models `conceptual emergence', while sector 2 of Fock space models a conceptual combination from the combining concepts by requiring 
the rules of logic for the logical connective used for the combining to be satisfied in a probabilistic setting. The relative prevalence of emergence  or logic in a specific cognitive process is measured by the `degree of participation' of sectors 1 and 2, respectively. The abundance of evidence of deviations from classical logical reasoning in concrete human decisions (paradoxes, fallacies, effects, contradictions), together with our results, led us to draw the conclusion that emergence constitutes the dominant dynamics of human reasoning, while logic is only a secondary form of dynamics. 

We now prove that the two-layered structure above explains and justifies the results and mathematics in Secs. \ref{newpattern} and \ref{quantum}.

Let us preliminarily observe that, if one reflects on how we represented conceptual negation in Sec. \ref{quantum}, one realizes at once that its modeling directly and naturally follows from the general assumption stated above. Indeed, suppose that a given subject is asked to estimate whether a given exemplar $x$ is a member of the concepts $A$, $B'$, `$A \ {\rm and} \ B'$ (a completely equivalent explanation can be given for the conjunctions `$A'$ and $B$' and `$A'$ and $B'$'). Then, our quantum mathematics can be interpreted by assuming that a `quantum logical thought' acts, where the subject considers two copies of $x$ and estimates whether the first copy belongs to $A$ and the second copy of $x$ `does not' belong to $B$. But also a `quantum emergent thought' acts, where the subject estimates whether the exemplar $x$ belongs to the newly emergent concept `$A \ {\rm and} \ B'$'. The place whether these superposed processes can be suitably structured is the two-sector Fock space. First sector of Fock space hosts the latter process, second sector hosts the former, and the weights $m_{AB'}^2$ and $n_{AB'}^2$ indicate whether the overall process is mainly guided by logic or emergence.

Let us now come to the significantly stable deviations from Eqs. (\ref{condbis01})--(\ref{condbis05}) in Sec. \ref{newpattern}. We have argued that these deviations occur at a different, deeper, level which could not be identified by only detecting overextension and underextension in combinations, and they are very likely to express a general mechanism determining how the human mind forms concepts. This would already be convincing even without mentioning a Fock space modeling. But, this very stable pattern can exactly be explained in our two-sector Fock space framework by assuming that emergence plays a primary role in the human reasoning process, but also aspects of logic are systematically present in human decision processes. We provide in the following a heuristic reasoning leading to this conclusion.

Suppose that only emergent reasoning is present in human thought, and represent it in sector 1 of Fock space. Consider, e.g., Eq. (\ref{condbis01}), and use Eq. (\ref{BornRuleSect1}) in it. We get
\begin{eqnarray}
&I_{A}=\mu(A)-\mu(A\ {\rm and}\ B)-\mu(A \ {\rm and}\ B')= \nonumber \\
&=\mu(A)-\frac{1}{2}(\mu(A)+\mu(B))-\Re \langle A|M|B\rangle-\frac{1}{2}(\mu(A)+\mu(B'))-\Re \langle A|M|B'\rangle= \nonumber \\
&= -\frac{1}{2}(\mu(B)+\mu(B'))- ( \Re \langle A|M |B\rangle+\Re \langle A|M |B'\rangle)
\end{eqnarray}
Let us now consider the interference terms and set $\Re \langle A|M|B\rangle={\cal I}_{AB}$ and $\Re \langle A|M|B'\rangle={\cal I}_{AB'}$. Further, let us consider the deviations of $\mu(B)$ and $\mu(B')$ from $\frac{1}{2}$, i.e. set ${\cal I}_{B}=\mu(B)-\frac{1}{2}$ and ${\cal I}_{B'}=\mu(B')-\frac{1}{2}$. This coincides with introducing a purely mathematical notion of `interference of an individual concept'. Then,
\begin{equation}
I_{A}=-\frac{1}{2}-\Big [ \frac{{\cal I}_{B}+{\cal I}_{B'}}{2}+{\cal I}_{AB}+{\cal I}_{AB'}  \Big ]
\end{equation}
Hence, the value of $I_A$ oscillates around the value -0.5 due to the interference terms ${\cal I}_{B}$, ${\cal I}_{B'}$, ${\cal I}_{AB}$ and ${\cal I}_{AB'}$. When one considers all exemplars, it is reasonable to assume that these interference terms offset each other, which entails $I_{A} \approx -0.5$. This assumption is confirmed by the experimental data. If one look at these data (see \cite{asv2015FRONTIERS}), one finds that the term $\frac{{\cal I}_{B}+{\cal I}_{B'}}{2}+{\cal I}_{AB}+{\cal I}_{AB'}$ is indeed
very small -- e.g., {\it Shelves} and {\it Fox} give -0.01, {\it Cinnamon} gives 0, {\it Blue-tit}, {\it Raising} and {\it Almond} give 0.01.

A conclusion follows at once. By assuming that only emergent reasoning is present, we get $I_{A} \approx -0.5$. One can use a similar heuristic reasoning to show that the same result holds for $I_B$, $I_A'$ and $I_{B'}$. Analogously, 
\begin{equation} 
I_{ABA'B'}=-1-\sum_{X=A,A'}\sum_{Y=B,B'}{\cal I}_{XY}-2 \Big (\sum_{X=A,A'} \frac{{\cal I}_{X}}{2}+\sum_{Y=B,B'} \frac{{\cal I}_{Y}}{2} \Big )
 \approx -1 
\end{equation}
where ${\cal I}_{Y}=\mu(Y)-\frac{1}{2}$, $Y=B,B'$. 

But, now, comparison with the experimental values of $I_{X}$, $I_{Y}$ and $I_{ABA'B'}$, $X=A,A'$, $Y=B,B'$, in Sec. \ref{newpattern} reveals that a component of sector 2 of Fock space is also present, which is generally smaller than the component of sector 1 but systematic across all exemplars. The consequence is immediate: both emergence and logic play a role in the decision process -- emergence is dominant, but also logic is systematically present.

Due to the nature of its appearance, we believe that this finding is deep and fundamental concerning the dynamics of conceptuality in human thought, and it deserves further investigation in the near future. But, already at the present stage of our research, we have good reasons to believe that our two-sector Fock space framework goes  essentially beyond faithful representation of one set or more sets of experimental data. It reveals fundamental aspects of the mechanisms through which concepts are formed, combine and constitute the substance of human thought.

\section{Conclusive remarks\label{conclusions}}
In this paper we have analyzed a set of experimental data we collected on combinations, i.e. conjunctions and negations, of two concepts. We have isolated five classicality conditions that should be satisfied in order to represent the collected data in a classical Kolmogorovian probability framework. These conditions were systematically violated in our experiments, but the usual overextension is not the only non-classical effect at play. Indeed, the deviation from the five classicality conditions we have identified is surprisingly independent of the specific exemplar that is considered, nor it depends on the specific pair of concepts that are measured. In other words, the numerical value of, say $I_A$, is the same when different exemplars are considered, and it is not affected by the specific pair of concepts that are measured. In addition, $I_A$ and $I_B$, and also $I_{A'}$ and $I_{B'}$, have approximately the same numerical value, and the slight deviation between their values can be attributed to the fact that a positive or a negative concept is considered. As a consequence, this deviation of classicality cannot be attributed to an underlying non-classical mechanism of conceptual combination, because the numerical value of the functions $I_A$, $I_B$, etc. would then depend on whether `$A$ and $B$', or `$A$ and $B'$', or `$A'$ and $B$', or `$A'$ and $B'$', is considered -- in $I_{A'}$ more negative concepts appear than in $I_A$. This is why, we think, this deeper phenomenon could not have been detected in a less symmetric experimental setting only aiming at detecting overextension, as we have explained in Sec. \ref{intro}. As it is reasonably independent of the specific conceptual combination that is considered, this non-classical effect unveils in our opinion aspects of the mechanisms of conceptual formation, and it can be numerically represented in our quantum-theoretic framework in two-sector Fock space. In addition, it can be explained by assuming that human reasoning is a specifically structured superposition of a `dominant emergent reasoning' and a `secondary logical reasoning', as we have argued in Sec. \ref{explanation}.

It is worth observing, to conclude, that our explanatory hypothesis on the existence of two supersposed processes in human reasoning  has appeared in different forms in cognitive psychology 
and is referred to as `dual route models of cognition'. Already at the beginning of the last century William James proposed a `dual process theory', where he introduced the idea of `two legs of thought', a `conceptual leg', being exclusive, static, classical and following the rules of logic, and a `perceptual leg', being intuitive and penetrating. He expressed the opinion that `just as we need two legs to walk, we also need both conceptual and perceptual modes to think' \cite{dual0}. Please, notice that James used the connotation `conceptual' to indicate the logical mode, contrary to us using `conceptual' mainly with respect to the emergent mode. Recently, more sophisticated dual models of cognition were put forward among the other by Sloman, who distinguished between `associative system of reasoning' and `rule based system of reasoning' \cite{dual1}, and by Evans, who proposed a theory of reasoning with `heuristic processes' and `analytic processes' \cite{dual2}. Notwithstanding their similarities, we believe there are some fundamental differences between earlier dual route models of cognition and ours. First of all, it is the specific mathematical structure of our quantum model that defines the structural aspects of the two layers of reasoning that we put forward here and how they are interrelated. In other words, the nature of this double layered structure follows from a mathematical model for experimental data on the non-classical effects identified in human cognition. Secondly, and directly related to the first difference, in our hypothesized double structure, conceptual reasoning and logical reasoning are superposed and entangled in a standard quantum-mechanical sense, rather than having an individual (or separated, or parallel) existence.

\newpage

\appendix

\section{Data tables\label{tables}}

\begin{table}[!h]
\caption{Values of $I_{A}$,  $I_{B}$,	$I_{A'}$, $I_{B'}$ and	$I_{ABA'B'}$ for the pair ({\it Home Furnishing}, {\it Furniture}).}
\label{table1}
\scriptsize
\begin{center}
\begin{tabular}{|c|c|c|c|c|c|}
\hline
\multicolumn{6}{|l|}{({\it Home Furnishing}, {\it Furniture})} \\
\hline		
Exemplar & $I_{A}$	& $I_{B}$	& $I_{A'}$	&	$I_{B'}$	&	$I_{ABA'B'}$	\\
\hline
{\it Mantelpiece}		&	-0.56	&	-0.31	&	-0.31	&	-0.46	&	-0.89	\\
{\it Window Seat}		&	-0.44	&	-0.36	&	-0.33	&	-0.35	&	-0.74	\\
{\it Painting}		&	-0.44	&	-0.48	&	-0.35	&	-0.33	&	-0.94	\\
{\it Light Fixture}		&	-0.48	&	-0.45	&	-0.33	&	-0.28	&	-0.84	\\
{\it Kitchen Counter}		&	-0.42	&	-0.44	&	-0.39	&	-0.24	&	-0.79	\\
{\it Bath Tub}		&	-0.45	&	-0.44	&	-0.37	&	-0.41	&	-0.83	\\
{\it Deck Chair}		&	-0.42	&	-0.38	&	-0.44	&	-0.39	&	-0.86	\\
{\it Shelves}		&	-0.38	&	-0.43	&	-0.36	&	-0.34	&	-0.83	\\
{\it Rug}		&	-0.48	&	-0.54	&	-0.45	&	-0.28	&	-1	\\
{\it Bed}		&	-0.39	&	-0.48	&	-0.49	&	-0.39	&	-0.9	\\
{\it Wall-Hangings}		&	-0.39	&	-0.44	&	-0.38	&	-0.27	&	-0.85	\\
{\it Space Rack}		&	-0.53	&	-0.41	&	-0.37	&	-0.44	&	-0.9	\\
{\it Ashtray}		&	-0.34	&	-0.45	&	-0.43	&	-0.35	&	-0.84	\\
{\it Bar}		&	-0.51	&	-0.39	&	-0.43	&	-0.51	&	-1.03	\\
{\it Lamp}		&	-0.51	&	-0.51	&	-0.45	&	-0.41	&	-1.05	\\
{\it Wall Mirror}		&	-0.58	&	-0.51	&	-0.45	&	-0.35	&	-1.06	\\
{\it Door Bell}		&	-0.39	&	-0.51	&	-0.53	&	-0.36	&	-0.99	\\
{\it Hammock}		&	-0.48	&	-0.5	&	-0.47	&	-0.41	&	-0.98	\\
{\it Desk}		&	-0.32	&	-0.58	&	-0.59	&	-0.39	&	-1	\\
{\it Refrigerator}		&	-0.47	&	-0.4	&	-0.46	&	-0.39	&	-0.93	\\
{\it Park Bench}		&	-0.31	&	-0.45	&	-0.36	&	-0.22	&	-0.79	\\
{\it Waste Paper Basket}		&	-0.31	&	-0.51	&	-0.59	&	-0.27	&	-0.95	\\
{\it Sculpture}		&	-0.48	&	-0.58	&	-0.49	&	-0.43	&	-1.13	\\
{\it Sink Unit}		&	-0.46	&	-0.41	&	-0.41	&	-0.36	&	-0.91	\\
\hline
\end{tabular}
\end{center}
\end{table}

\begin{table}[!h]
\caption{Values of $I_{A}$,  $I_{B}$,	$I_{A'}$, $I_{B'}$ and	$I_{ABA'B'}$ for the pair ({\it Spices}, {\it Herbs}).}
\label{table2}
\scriptsize
\begin{center}
\begin{tabular}{|c|c|c|c|c|c|}
\hline
\multicolumn{6}{|l|}{({\it Spices}, {\it Herbs})} \\
\hline		
Exemplar & $I_{A}$	& $I_{B}$	& $I_{A'}$	&	$I_{B'}$	&	$I_{ABA'B'}$	\\
\hline
{\it Molasses}		&	-0.41	&	-0.36	&	-0.31	&	-0.43	&	-0.75	\\
{\it Salt}		&	-0.26	&	-0.28	&	-0.33	&	-0.37	&	-0.61	\\
{\it Peppermint}		&	-0.41	&	-0.33	&	-0.33	&	-0.43	&	-0.78	\\
{\it Curry}		&	-0.45	&	-0.42	&	-0.34	&	-0.31	&	-0.79	\\
{\it Oregano}		&	-0.38	&	-0.43	&	-0.36	&	-0.35	&	-0.76	\\
{\it MSG	}	&	-0.36	&	-0.34	&	-0.37	&	-0.45	&	-0.76	\\
{\it Chili Pepper}		&	-0.73	&	-0.54	&	-0.35	&	-0.46	&	-1.1	\\
{\it Mustard}		&	-0.49	&	-0.44	&	-0.3	&	-0.41	&	-0.83	\\
{\it Mint}		&	-0.46	&	-0.47	&	-0.32	&	-0.34	&	-0.85	\\
{\it Cinnamon}		&	-0.48	&	-0.41	&	-0.34	&	-0.43	&	-0.84	\\
{\it Parsley}		&	-0.4	&	-0.5	&	-0.36	&	-0.35	&	-0.84	\\
{\it Saccarin}		&	-0.43	&	-0.34	&	-0.36	&	-0.46	&	-0.81	\\
{\it Poppy Seeds}		&	-0.43	&	-0.43	&	-0.29	&	-0.4	&	-0.84	\\
{\it Pepper}		&	-0.61	&	-0.41	&	-0.21	&	-0.46	&	-0.91	\\
{\it Turmeric}		&	-0.54	&	-0.49	&	-0.38	&	-0.47	&	-0.91	\\
{\it Sugar}		&	-0.46	&	-0.26	&	-0.31	&	-0.44	&	-0.81	\\
{\it Vinegar}		&	-0.26	&	-0.31	&	-0.33	&	-0.36	&	-0.65	\\
{\it Sesame Seeds}		&	-0.49	&	-0.44	&	-0.33	&	-0.4	&	-0.91	\\
{\it Lemon Juice}		&	-0.3	&	-0.34	&	-0.46	&	-0.43	&	-0.78	\\
{\it Chocolate}		&	-0.39	&	-0.36	&	-0.37	&	-0.44	&	-0.81	\\
{\it Horseradish}		&	-0.4	&	-0.47	&	-0.37	&	-0.44	&	-0.86	\\
{\it Vanilla}		&	-0.48	&	-0.44	&	-0.38	&	-0.48	&	-0.91	\\
{\it Chives}		&	-0.38	&	-0.51	&	-0.53	&	-0.33	&	-0.99	\\
{\it Root Ginger}		&	-0.43	&	-0.54	&	-0.41	&	-0.37	&	-0.91	\\
\hline
\end{tabular}
\end{center}
\end{table}

\begin{table}[!h]
\caption{Values of $I_{A}$,  $I_{B}$,	$I_{A'}$, $I_{B'}$ and	$I_{ABA'B'}$ for the pair ({\it Pets}, {\it Farmyard Animals}).}
\label{table3}
\scriptsize
\begin{center}
\begin{tabular}{|c|c|c|c|c|c|}
\hline
\multicolumn{6}{|l|}{({\it Pets}, {\it Farmyard Animals})} \\
\hline		
Exemplar & $I_{A}$	& $I_{B}$	& $I_{A'}$	&	$I_{B'}$	&	$I_{ABA'B'}$	\\
\hline
{\it Goldfish}		&	-0.41	&	-0.43	&	-0.48	&	-0.53	&	-0.94	\\
{\it Robin}		&	-0.39	&	-0.41	&	-0.22	&	-0.18	&	-0.59	\\
{\it Blue-tit}		&	-0.31	&	-0.3	&	-0.24	&	-0.24	&	-0.56	\\
{\it Collie Dog}		&	-0.48	&	-0.34	&	-0.34	&	-0.33	&	-0.79	\\
{\it Camel}		&	-0.36	&	-0.46	&	-0.3	&	-0.24	&	-0.7	\\
{\it Squirrel}		&	-0.24	&	-0.34	&	-0.31	&	-0.2	&	-0.59	\\
{\it Guide Dog for Blind}		&	-0.35	&	-0.39	&	-0.36	&	-0.36	&	-0.76	\\
{\it Spider}		&	-0.31	&	-0.36	&	-0.23	&	-0.19	&	-0.58	\\
{\it Homing Pigeon}		&	-0.41	&	-0.44	&	-0.31	&	-0.25	&	-0.74	\\
{\it Monkey}		&	-0.29	&	-0.31	&	-0.25	&	-0.31	&	-0.59	\\
{\it Circus Horse}		&	-0.39	&	-0.38	&	-0.26	&	-0.23	&	-0.69	\\
{\it Prize Bull}		&	-0.57	&	-0.49	&	-0.28	&	-0.35	&	-0.86	\\
{\it Rat}		&	-0.29	&	-0.39	&	-0.31	&	-0.23	&	-0.65	\\
{\it Badger}		&	-0.24	&	-0.3	&	-0.23	&	-0.19	&	-0.5	\\
{\it Siamese Cat}		&	-0.5	&	-0.41	&	-0.36	&	-0.46	&	-0.9	\\
{\it Race Horse}		&	-0.54	&	-0.46	&	-0.26	&	-0.24	&	-0.79	\\
{\it Fox}		&	-0.33	&	-0.34	&	-0.19	&	-0.19	&	-0.51	\\
{\it Donkey}		&	-0.45	&	-0.48	&	-0.26	&	-0.25	&	-0.78	\\
{\it Field Mouse}		&	-0.3	&	-0.24	&	-0.18	&	-0.23	&	-0.46	\\
{\it Ginger Tom-cat}		&	-0.34	&	-0.34	&	-0.34	&	-0.32	&	-0.71	\\
{\it Husky in Slead team}		&	-0.43	&	-0.49	&	-0.36	&	-0.28	&	-0.8	\\
{\it Cart Horse}		&	-0.46	&	-0.5	&	-0.31	&	-0.28	&	-0.79	\\
{\it Chicken}		&	-0.46	&	-0.44	&	-0.19	&	-0.23	&	-0.68	\\
{\it Doberman Guard Dog}		&	-0.47	&	-0.49	&	-0.54	&	-0.51	&	-1.03	\\
\hline
\end{tabular}
\end{center}
\end{table}

\begin{table}[!h]
\caption{Values of $I_{A}$,  $I_{B}$,	$I_{A'}$, $I_{B'}$ and	$I_{ABA'B'}$ for the pair ({\it Fruits}, {\it Vegetables}).}
\label{table4}
\scriptsize
\begin{center}
\begin{tabular}{|c|c|c|c|c|c|}
\hline
\multicolumn{6}{|l|}{({\it Fruits}, {\it Vegetables})} \\
\hline		
Exemplar & $I_{A}$	& $I_{B}$	& $I_{A'}$	&	$I_{B'}$	&	$I_{ABA'B'}$	\\
\hline
{\it Apple}		&	-0.49	&	-0.5	&	-0.3	&	-0.24	&	-0.79	\\
{\it Parsley}		&	-0.53	&	-0.51	&	-0.29	&	-0.29	&	-0.83	\\
{\it Olive	}	&	-0.46	&	-0.53	&	-0.41	&	-0.26	&	-0.86	\\
{\it Chili Pepper}		&	-0.53	&	-0.46	&	-0.29	&	-0.29	&	-0.83	\\
{\it Broccoli}		&	-0.58	&	-0.49	&	-0.21	&	-0.28	&	-0.83	\\
{\it Root Ginger}		&	-0.46	&	-0.46	&	-0.33	&	-0.24	&	-0.74	\\
{\it Pumpkin}		&	-0.43	&	-0.51	&	-0.29	&	-0.13	&	-0.68	\\
{\it Raisin}		&	-0.39	&	-0.51	&	-0.46	&	-0.33	&	-0.86	\\
{\it Acorn}		&	-0.36	&	-0.44	&	-0.39	&	-0.36	&	-0.84	\\
{\it Mustard}		&	-0.44	&	-0.45	&	-0.43	&	-0.38	&	-0.81	\\
{\it Rice}		&	-0.32	&	-0.34	&	-0.28	&	-0.29	&	-0.61	\\
{\it Tomato}		&	-0.56	&	-0.55	&	-0.33	&	-0.24	&	-0.86	\\
{\it Coconut}		&	-0.33	&	-0.44	&	-0.37	&	-0.33	&	-0.79	\\
{\it Mushroom}		&	-0.33	&	-0.33	&	-0.26	&	-0.24	&	-0.61	\\
{\it Wheat}		&	-0.38	&	-0.44	&	-0.38	&	-0.26	&	-0.73	\\
{\it Green Pepper}		&	-0.5	&	-0.49	&	-0.23	&	-0.26	&	-0.76	\\
{\it Watercress}		&	-0.45	&	-0.51	&	-0.24	&	-0.2	&	-0.73	\\
{\it Peanut}		&	-0.41	&	-0.43	&	-0.3	&	-0.33	&	-0.8	\\
{\it Black Pepper}		&	-0.38	&	-0.46	&	-0.31	&	-0.23	&	-0.71	\\
{\it Garlic}		&	-0.5	&	-0.49	&	-0.33	&	-0.31	&	-0.83	\\
{\it Yam}		&	-0.45	&	-0.58	&	-0.38	&	-0.24	&	-0.91	\\
{\it Elderberry}		&	-0.36	&	-0.52	&	-0.39	&	-0.28	&	-0.8	\\
{\it Almond}		&	-0.33	&	-0.42	&	-0.43	&	-0.37	&	-0.8	\\
{\it Lentils}		&	-0.38	&	-0.41	&	-0.33	&	-0.26	&	-0.71	\\
\hline
\end{tabular}
\end{center}
\end{table}

\cleardoublepage

\begin{table}[!h]
\caption{Correlation matrix for $I_{A}$,  $I_{B}$,	$I_{A'}$, $I_{B'}$ and	$I_{ABA'B'}$.}
\label{table5}
\scriptsize
\begin{center}
\begin{tabular}{|c|c|c|c|c|c|}
\hline		
& $I_A$ &	$I_B$ &	$I_{A'}$ &	$I_B'$ & $I_{ABA'B'}$ \\
\hline
$I_A$ &	1 &	0.45 &	-0.03 &	0.33 &	0.61 \\
$I_B$ &	0.45	& 1	& 0.41 & -0.07 &	0.63 \\
$I_{A'}$ &	-0.03 &	0.41	& 1	& 0.48 &	0.71 \\
$I_B'$	& 0.33 &	-0.07& 	0.48 &	1	& 0.61 \\
$I_{ABA'B'}$ &	0.61 &	0.63 &	0.71 &	0.61 &	1 \\
\hline
\end{tabular}
\end{center}
\end{table}

\end{document}